# Enhancing Preference Detection using Eye Movement Cascade and Multilayer Combinatorial Fusion

**Yuanhong Wu[†], Mary Tsaryk[†], Sam Watson[†], Tahsin Ahmed[†], Christina Schweikert[‡], D. Frank Hsu[†]**

**[†]Laboratory of Informatics and Data Mining, Department of Computer and Information Science, Fordham University, New York, NY, USA**

**[‡]Division of Computer Science, Mathematics and Science, St. John's University, Queens, NY, USA**

## ABSTRACT

*Understanding how visual attention influences human preference formation is a fundamental problem in consumer behavior and cognitive science. In this work, we investigate how eye movement patterns can be used to predict preference decisions when subjects compare two alternatives. To address this problem, we analyze an eye-tracking dataset, where subjects view pairs of face images and their gaze behavior is recorded throughout the decision process. We model this problem using multiple machine learning classifiers and apply the Multi-layer Combinatorial Fusion (MCF) framework to iteratively combine diverse models. The MCF framework expands models to a large pool of model fusions and then reduces them based on cognitive diversity, repeating this process across layers until diversity diminishes. Experimental results show that MCF significantly improves performance over individual models. The best single model achieves a Pre@k of 0.90, which is improved to 0.9143 after the first layer of MCF, and further increased to 0.9286 in the second layer. This progression demonstrates that the second layer not only surpasses the performance of the first layer but also outperforms the best individual model, highlighting the effectiveness of leveraging diversity through multi-layer fusion.*

## INTRODUCTION

In everyday consumer environments such as supermarkets, retail stores, and online shopping platforms, individuals are constantly exposed to multiple competing products and must make rapid preference decisions. When browsing items on shelves or screens, consumers visually compare alternatives before selecting one to purchase. These decisions are often formed within seconds and are influenced by both conscious evaluation and unconscious cognitive processes (Milosavljevic et al., 2011; Martin & Morich, 2011). Understanding how consumers visually interact with products and how these interactions translate into preferences is a fundamental problem in consumer behavior research (Pieters & Warlop, 1999; Orquin & Loose, 2013).

Traditional approaches to studying consumer preferences often rely on stated preference data collected through surveys, ratings, intentions, satisfaction measures, and conjoint or choice-based tasks, as well as revealed behavioral data such as purchase histories, transaction records (Rossi et al., 1996; Oard & Kim, 1998; Wedel & Kannan, 2016; Ben-Akiva, 2019; Krivulin & Gladkikh, 2022). While useful, these methods have limited access to the real-time cognitive processes that occur during decision making (Glaholt & Reingold, 2011; Zuschke, 2020; Wedel et al., 2023). In contrast, eye movement analysis provides a direct and continuous measure of visual attention, offering insights into how preferences evolve over time. By analyzing gaze patterns—such as fixation duration, gaze frequency, and transitions between items—we can better understand how consumers evaluate alternatives and form choices.

The importance of modeling consumer preference extends across multiple domains, including marketing, product placement, user interface design, and recommendation systems (Schafer et al., 2001; Dew et al., 2024). For example, understanding which visual features attract attention can help optimize product positioning on store shelves or digital interfaces (Chandon et al., 2009; Bialkova et al., 2020). Moreover, identifying the relationship between attention and preference can improve predictive models of consumer choice, enabling more personalized and effective decision support systems (Imai et al., 2019). Despite its importance, accurately capturing and modeling preference formation remains challenging due to its dynamic and context-dependent nature (Bettman et al., 1998; Otto et al., 2022).

A substantial body of research in psychology, marketing, and neuroscience has investigated how preferences are formed and how visual attention contributes to decision making. Early studies have shown that factors such as familiarity, novelty, and contextual presentation can significantly influence consumer choices (Zajonc, 1968; Berlyne, 1970; Tversky & Simonson, 1993). More recent work has focused on the role of attention in guiding decisions, suggesting that what individuals look at—and for how long—can both reflect and shape their preferences (Shimojo et al, 2003; Orquin & Loose, 2013). These findings motivate the use of behavioral signals, such as eye movements, as a proxy for underlying cognitive processes.

In the field of eye movement research, the relationship between gaze behavior and preference formation has been extensively studied. In particular, the pioneering work by Shimojo et al. (2003) introduced the gaze cascade effect, demonstrating that when subjects are presented with two alternatives, their gaze gradually shifts toward the option they eventually choose. This phenomenon indicates that visual attention is not merely passive observation but plays an active role in shaping decisions. Subsequent studies have shown that features such as fixation duration, gaze count, and temporal viewing patterns are useful predictors of preference outcomes (Glaholt, et al., 2009; Atalay et al., 2012; Gere et al., 2021).

Multi-layer combinatorial fusion (MCF), proposed by Hurley et al. (2020), is a fusion framework designed to operate across multiple layers. It extends the single-layer framework of combinatorial fusion analysis (CFA), introduced by Hsu et al. (2006, 2024), which provides systematic methods for combining multiple scoring systems. CFA has been applied in a wide range of domains, including virtual screening and drug discovery (Jiang et al., 2023), text classification (Xu et al., 2025), cybersecurity DoS detection (Alfatemi et al., 2025), and Bitcoin price prediction (Wu et al., 2025), value alignment of LLMs (Wu et al., 2026). MCF is an emerging technique that iteratively performs expansion and reduction through model combination, allowing it to explore additional performance gains beyond those achievable by single-layer CFA. This paper is one of the first applications of MCF to preference detection.

In this study, we utilize the eye movement dataset developed by Shimojo et al. (2011) to study preference detection through the MCF framework. Several prior studies have explored this dataset (Schweikert et al., 2016; Schweikert et al., 2018; Schweikert et al., 2021; Schweikert et al., 2022; Zhang et al., 2023). For example, Schweikert et al. (2016, 2021) applied the CFA framework to combine multiple features and demonstrated that CFA-based combinations can outperform the best individual feature. In subsequent work, Schweikert et al. (2022) further analyzed the diversity among the 12 subjects by constructing a prototypical subject and measuring differences using cognitive diversity. Building on these studies, we train five diverse machine learning models on the dataset and apply MCF to iteratively combine them across multiple layers. The fusion process continues until the diversity among the models becomes negligible. Experimental results show that MCF progresses through two layers and terminates at the third layer. The best performance achieved at the second layer surpasses both the first-layer results and the performance of the best original model. These findings demonstrate that MCF is an effective fusion framework for leveraging model diversity to improve predictive performance.

The remainder of this paper is organized as follows. Section 2 introduces the methods used for preference detection, including combinatorial fusion analysis (CFA) and multi-layer combinatorial fusion (MCF). Section 3 describes the eye movement dataset and the training procedure for the five machine learning models. Section 4 presents the experimental results of the MCF framework. Section 5 concludes the paper.

# METHODOLOGY

This section introduces the combinatorial fusion method used to analyze the eye movement data, including combinatorial fusion analysis (CFA) and multi-layer combinatorial fusion (MCF). CFA combines models within a single layer, while MCF extends this approach by iteratively integrating combinations across multiple layers.

## Combinatorial Fusion Analysis

**Rank-score function.** CFA operates on scoring systems. A scoring system is a broad concept and appears widely in everyday life. Any system that assigns numerical values to observations can be considered as a scoring system. For example, a thermometer assigns a temperature value at each point in time, and a blood pressure monitor produces measurements for each patient, etc.

A scoring system in CFA consists of three components: a score function, a rank function and a rank-score function. The score function represents the original numerical values assigned by the system. The rank function is obtained by ordering these scores in descending order and assigning ranks accordingly. The rank-score function combines both the score and rank information. Specifically, let $D = \{d_1, d_2, \cdots, d_n\}$ denotes a set of data items, where each $d_i$ is an observation, and let $A$ be a scoring system defined on $D$. The score function of $A$, denoted by $s_A$, maps each data item to a real value, i.e., $s_A: D \to R$. By sorting these scores in descending order, we obtain rank function $r_A: D \to N$, where rank 1 corresponds to the highest score and rank $n$ to the lowest. Then rank-score function $f_A$ is defined by combining $s_A$ and $r_A$ for each data item, as follows (Hsu et al. 2006; Hsu et al. 2010):

$$f_A = s_A(\, r_A^{-1}(i)\,) = (s_A \circ r_A^{-1})(i), \qquad i \in D.$$

The relationship among score function, rank function and rank-score function is depicted in Figure 1. We note that these three function are interconnected. Given any two of them, the third can be derived.

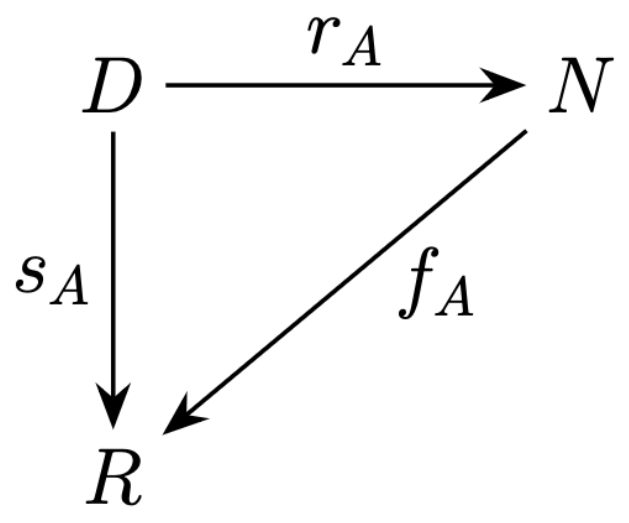

**Fig. 1:** Relationship among score function $s_A$, rank function $r_A$ and rank-score function $f_A$ (Hsu et al., 2010).

**Cognitive diversity and diversity strength.** Cognitive diversity, introduced in (Hsu et al., 2010; Hsu et al., 2019), measures the dissimilarity between two scoring systems. Unlike conventional statistical correlation measures such as Pearson's correlation, Spearman's correlation, and Kendall's tau correlation, as well as computation diversity measures like entropy and KW variance (Hurley et al., 2021), cognitive diversity is designed to captures the differences in the rank-score behavior of scoring systems. Let A and B be two scoring systems with rank-score functions $f_A$ and $f_B$ defined on data items $D$. The cognitive diversity between A and B is defined as (Jiang et al., 2023):

$$CD(A, B) = \sqrt{\tfrac{1}{n}\textstyle\sum_{i=1}^{n}(\, f_A(d_i) - f_B(d_i)\,)^2} \qquad i \in D.$$

Cognitive diversity is a pairwise diversity measure. In order to use it as a fusion weight, we need to extend it to a system-level diversity measure, which leads to a notion of diversity strength. It quantifies the global diversity of a given system relative to all other systems in a system set. Specifically, for a set of $t$ scoring systems, the diversity strength of system $A_i$ is computed by averaging its cognitive diversity with respect to all other systems in the set. The formula of the diversity strength of system $A_i$ is given as follows (Jiang et al., 2023):

$$DS(A_i) = \frac{1}{t-1}\sum_{\substack{j=1,\\ j\neq i}}^{t} CD(A_i, A_j), \qquad i,j \in D.$$

**Four combination strategies.** CFA generates a large number of model combinations through four combination strategies. Because each scoring system includes both a score function and a rank function, models can be combined using their scores or ranks. For combination weights, we consider two approaches: equal weighting and weighting by diversity strength. By combining these two dimensions—combination method and combination weight—we obtain four types of model combinations. Table 1 summarizes these four combination types.

| Combination weight | Combination method | |
|---|---|---|
| | score | rank |
| Equal weight | ASC | ARC |
| diversity strength | WSCDS | WRCDS |

**Table 1**: Summary of four combination strategies in CFA, where ASC and ARC represent average score and rank combination, while WSCDS and WRCDS are weighted score and rank combination by diversity strength.

The corresponding formulas for these four types of combinations are given below (Jiang et al., 2023):

$$s_{ASC}(d_i) = \frac{1}{t}\sum_{j=1}^{t} s_{A_i}(d_i), \qquad s_{ARC}(d_i) = \frac{1}{t}\sum_{j=1}^{t} r_{A_i}(d_i),$$

$$s_{WSCDS}(d_i) = \sum_{j=1}^{t} \frac{DS(A_j)}{\sum_{k=1}^{t} DS(A_j)} s_{A_i}(d_i), \qquad s_{WRCDS}(d_i) = \sum_{j=1}^{t} \frac{1/DS(A_j)}{\sum_{k=1}^{t} 1/DS(A_j)} r_{A_i}(d_i).$$

In CFA, for each combination strategies, the combination size can vary from 2 to $t$, which results in $\binom{t}{2} + \binom{t}{3} + \cdots + \binom{t}{t} = 2^t - 1 - t$ combined models. Therefore, across all four combination strategies, the full CFA framework can generate in total $4 \times (2^t - 1 - t)$ model combinations.

## Multi-layer Combinatorial Fusion

**EAR algorithm.** The MCF framework performs iterative model fusion through an Expansion-and-Reduction (EAR) algorithm (Hurley et al., 2021). It begins with a set of $t$ diverse scoring systems. The framework first expands them to $4 \times (2^t - 1 - t)$ combinations by applying the six fusion strategies in CFA. Then this large pool of combined models is reduced back to $t$ models using a reduction method. In this study, diversity-based on reduction method is used. Specifically, cognitive diversity is calculated for each pair of combined models and a diversity strength is derived for each individual model to reflect its overall distinctiveness within the set. These diversity strengths are then sorted in descending order. And then the top $t$ models with the highest diversity strength are selected as the base models for the next layer of the MCF framework.

MCF proceeds by repeatedly applying expansion and reduction steps as it moves to deeper layers. At each layer, we visualize the base models using the rank-score function graph and assess their diversity. When the graph indicates that the models exhibit little to no diversity, the process terminates. In this study, when the maximum diversity strength among the base models falls below 0.05, we stop the MCF iteration. This is because MCF is diversity-driven framework. When diversity is negligible, fusions barely help.

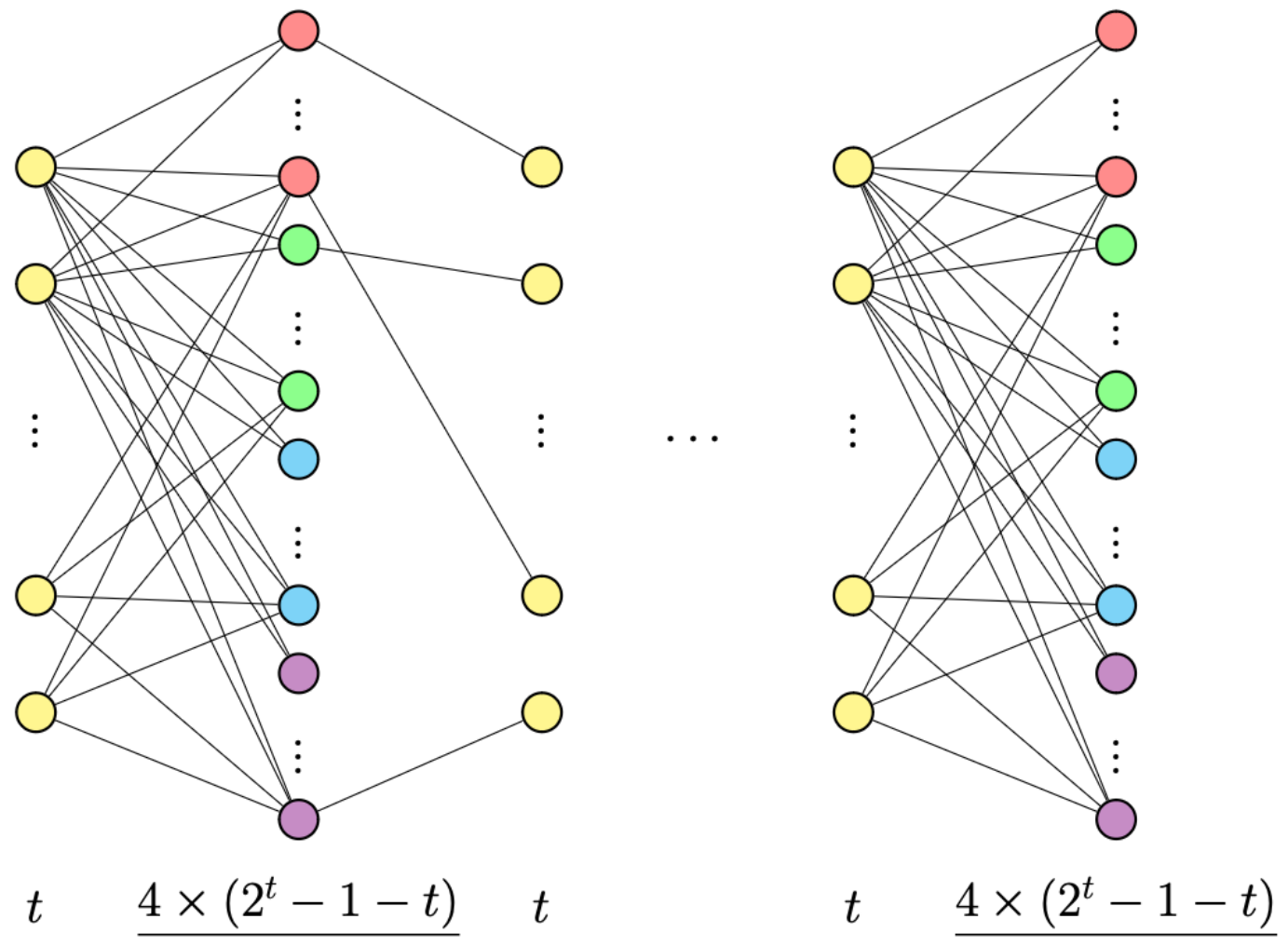

**Fig 2:** The MCF framework with $t$ initial base models. The red, green, blue, and purple nodes in the expansion step correspond to the four types of combinations.

**MCF framework.** The diagram of the MCF framework with $t$ initial base models is shown in Figure 2. In each layer, the $t$ models are expanded to $4 \times (2^t - 1 - t)$ combined models (4 node colors represent 4 types of combinations; each color has $2^t - 1 - t$ nodes), after which the top $t$ most diverse models based on diversity strength are selected to serve as the input for the next layer, where the expansion and reduction is repeated. This process continues until there is almost no diversity among the base models. Finally, the $4 \times (2^t - 1 - t)$ models on the last layer are evaluated with the target variable.

# EXPERIMENTS

## Dataset description

The eye movement dataset used in this study comes from the preference decision-making experiment conducted by Shimojo et al (2011). Their experiment was designed to investigate how gaze behavior is related to preference information. In each trial, a subject was shown two computer-generated face images, one on the left and one on the right. And they were asked to choose the face they considered more attractive. During this process, the subject's eye movements were recorded, including $x, y$ coordinates of gaze points and the fixation duration at each point. Each subject completed 60 trials, and the final preference decision for each trial was also stored. Figure 3 shows an example trial plot with the subject's gaze points. The selected face image is highlighted in green.

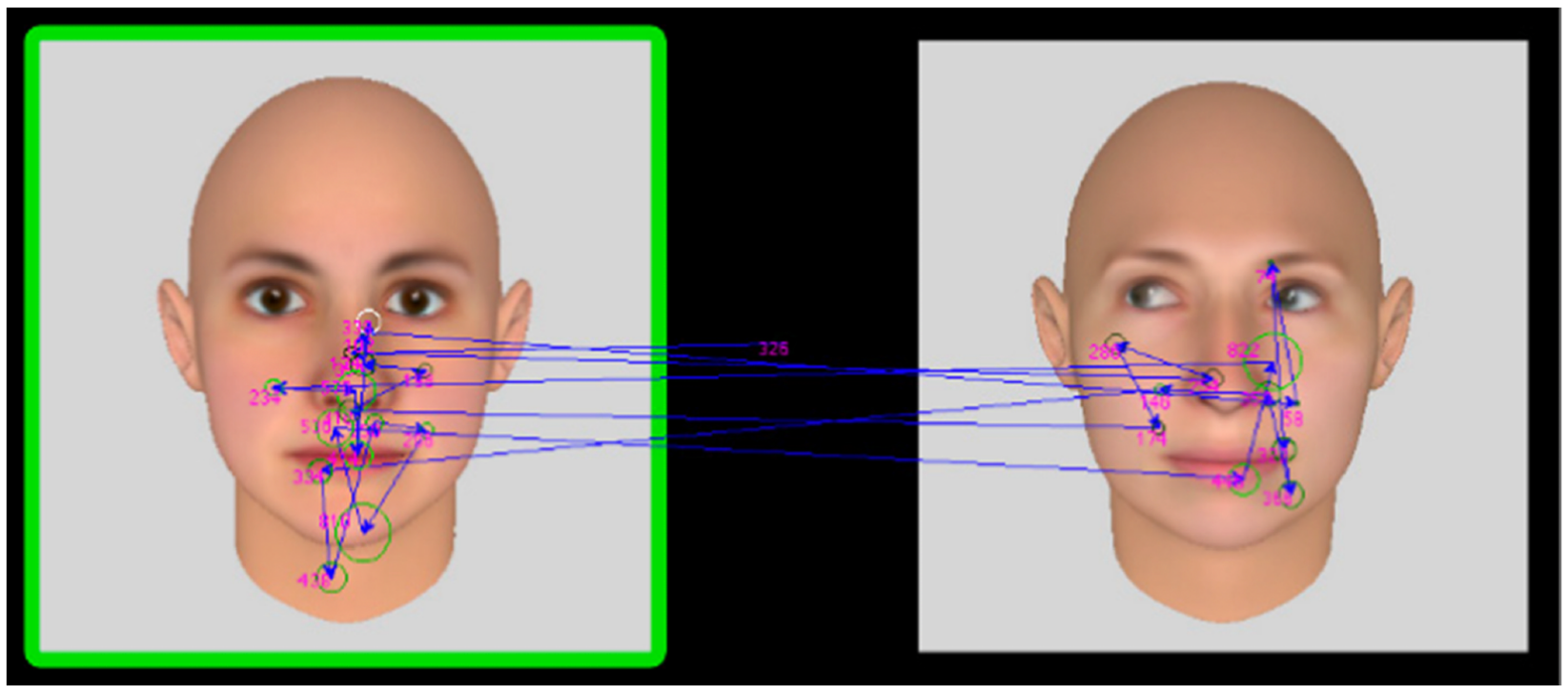

**Figure 3:** A sample trial plot with the subject's gaze points (Shimojo et al., 2011).

Based on these recorded gaze sequences, five features are constructed to characterize visual attention and decision behavior. These features include the time spent on each face in the last 200ms of a trial, the total time spent on each face throughout the trail, the number of gaze fixation points on each face, the sustainability of visual interest over time, and the frequency of gaze transitions among facial regions such as the eyes, nose, and mouth. These features allow us to analyze how attention is distributed the between two face images and how gaze patterns affect the final preference choice. Each feature is defined as the difference between its value for the left face and the corresponding value for the right face. For each subject, the feature values across all trials are then normalized to the range [0, 1] using min-max normalization.

## Base models and training procedure

In this study, machine learning models are used to construct scoring systems. Based on combinatorial fusion theory and prior empirical practice, combining a small set of models that are diverse and accurate can improve upon the best single model. In practice, CFA usually fuses a limited number of models, such as four, five or six, rather than a large number of models. Although adding more models may further improve performance, the marginal gain often diminishes as the ensemble size increases, while the computational cost continues to grow. Therefore, model selection involves a trade-off between potential performance gain and computational complexity.

In this paper, we use five machine learning models: Gaussian Naïve Bayes (A), Decision Tree (B), Random Forest (C), Gradient Boosting (D), and Multilayer Perceptron (E). These models are selected because they represent distinct learning principles and model architectures, which can lead to diverse scoring behaviors. Gaussian Naïve Bayes is probabilistic Bayesian classifier that estimates class posterior probabilities under a conditional independence assumption among features. This makes it fundamentally different from discriminative models, as it models the probability structure of the data rather than directly learning decision boundaries. Decision Tree is a rule-based model that recursively partitions the feature space using hierarchical decision rules. Random Forest builds a parallel ensemble of decision trees which randomly selects features and samples. Gradient Boosting is also tree-based, but it builds trees sequentially, with each new tree focusing on correcting the errors made by previous trees. In contrast, Multilayer Perceptron is a neural network model that learns nonlinear feature representations through multiple layers of weighted transformations and activation functions. This architectural and algorithmic diversity across models provides the basis for combinatorial fusion to exploit complementary information.

The eye movement dataset contains no missing values, and all features are within the same range [0, 1]. Therefore, no additional preprocessing is required before model training. The data are partitioned into training and test sets with a ratio of 80:20. On the training set, 3-fold stratified cross-validation is employed for each model to ensure robust performance estimation while preserving class distribution. The test set is held out for the final evaluation. After training, each model generates probabilistic predictions on the test set, which are scoring systems.

Following prior studies on the eye movement dataset (Schweikert et al., 2022; Zhang et al., 2023), we evaluate model performance using precision at k (Pre@$k$). Pre@$k$ is a rank-based evaluation metric that measures the proportion of truly positive instances among the top $k$ ranked items. This metric is widely used in information retrieval, recommender systems, and other rank-based decision tasks (Manning 2008; Cremonesi et al. 2010). In our test set, there are 144 data items, including 71 positive instances. Therefore, we set $k = 70$, which approximately matches the number of positive samples in the test set. An ideal model would rank all positive instances before any negative instances, thereby achieving a high Pre@$k$ score. Pre@$k$ is computed as the number of true positive instances among the top 70 data items divided by 70.

## Fusion of scoring systems

The scoring systems are integrated using combinatorial fusion, which consists of CFA and MCF. CFA serves as the foundation of MCF. In this study, five base models were used; therefore, ($t$=5) in Figure 2. CFA combines these five models using four types of fusion strategies, defined by two input forms—score and rank—and two weighting schemes—simple averaging and diversity-strength-based weighting. For score-based fusion, the original predictive probabilities are directly used. For rank-based fusion, the predictive probabilities are first sorted in descending order and converted into ranks. Through CFA expansion, the five original models are expanded into $4 \times (2^5 - 1 - 5) = 104$ fused models. A reduction step is then applied to prune these 104 models back to five models based on diversity strength. Together, the CFA-based expansion and diversity-strength-based reduction constitute one iteration of the MCF framework. This process can be repeated until the diversity strength falls below the predefined threshold of 0.05.

# RESULTS

In this section, we present results of the MCF framework on the eye movement dataset.
The MCF framework proceeds iteratively over three layers. Figure 4 illustrates the evolution of rank-score function graphs throughout this process. In the first layer, the five models are relatively well separated, indicating a moderate level of diversity among their scoring behaviors. As the framework advances to the second layer, all models, except model B, becomes closely aligned, suggesting a big reduction in diversity, while model B exhibits distinct behavior compared to other four models. In the third layer, the five models nearly overlap, which shows that the diversity among models have reduced to almost zero. At this point, further fusion is unlikely to yield meaningful improvement. The MCF process stops at the third layer. Overall, these observations demonstrate that model diversity decreases gradually across layers. The MCF leverages this diversity in each layer to achieve performance gains.

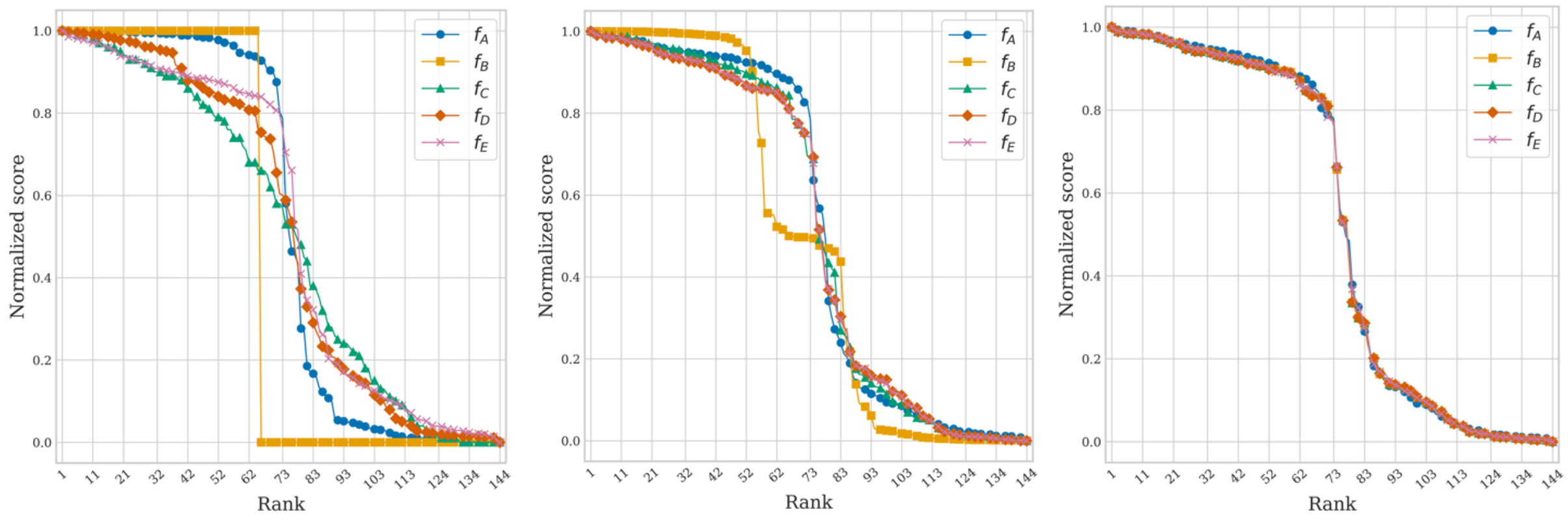

**Fig 4:** Rank-score function graphs for layer 1 (left), layer 2 (middle) and layer 3 (right).

As shown in the rank-score function graph of the first layer in Figure 4, model B exhibits a large number of ties. The presence of ties introduces ambiguity in rank combinations, since multiple data items are assigned the same rank, which reduces the resolution of the ranking and obscures fine-grained distinctions between data items. When this ambiguity is propagated into rank-based fusion, it can degrade the quality of rank combination and introduce instability. Moreover, in the iterative MCF framework, these inaccuracies may accumulate across layers, further diminishing the effectiveness of subsequent combinations. Therefore, we restrict the first layer to score-based fusion only. We consider two types of score combination in the first layer: average score combination (ASC) and weighted score combination by diversity strength (WSCDS).

After evaluating all combined models, the performance results are displayed in Figure 5. The models in x axis are organized into five distinct blocks based on combination size (ranging from single base models to 5-model combinations). The y axis represents the performance metric Pre@$k$, and the horizontal red dashed line indicates the best performance achieved by any individual model, which is 0.9. In the figure, purple crosses correspond to ASC models, while blue triangles represent WSCDS models.

From the figure, it's evident that combinations from ASC generally achieve performance than those from WSCDS in the first layer. Furthermore, two combined models outperform the best individual model: the ADE combination under ASC and the ACDE combination under WSCDS. Both achieve a Pre@$k$ of 0.9143, corresponding to 1.43% improvement over the strongest single model. This result shows the effectiveness of MCF at the first layer.

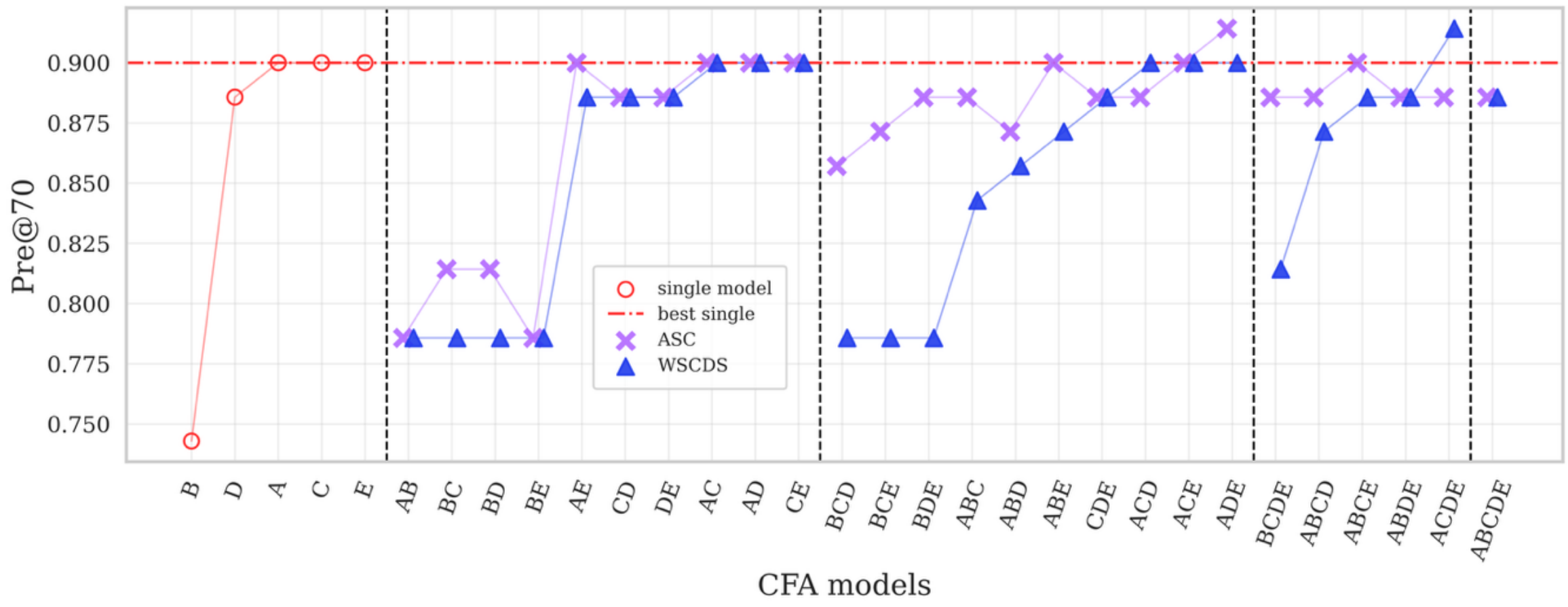

**Fig. 5:** Performance plot for score combination (ASC, WSCDS) in the first layer.

In the first layer, the initial 5 models are expanded into 52 combined models through score combination, which are then pruned back to 5 models by selecting those with the highest diversity strength. These refined models are the base models for the second layer. Since the rank-score function graph in the second layer show no ties, we apply both score combination and rank combination. As a result, the second layer produces 104 combined models, consisting of 52 models from score combination and 52 models from rank combination. The performance results are presented in two separate plots: one for average combinations (ASC and ARC) and the other for weighted combinations (WSCDS and WRCDS), which provides clearer comparison between score-based and rank-based fusion strategies.

Figure 6 illustrates the performance of average combinations within the second layer. The top-performing individual models achieve a Pre@$k$ of 0.9. When combining two models, two rank-based combinations reach a performance of 0.9143, showing an improvement over the best base model. As the number of combined models increases, we observe further gains. In particular, the combinations ABD, ABE, and ABCDE under average score combination, as well as BCDE and ABCDE under average rank combination, achieve a Pre@$k$ of 0.9286. This corresponds to a 2.86% improvement over the best single model at this layer. These results demonstrate that both score-based and rank-based average combinations are very effective, leading to consistent performance gains.

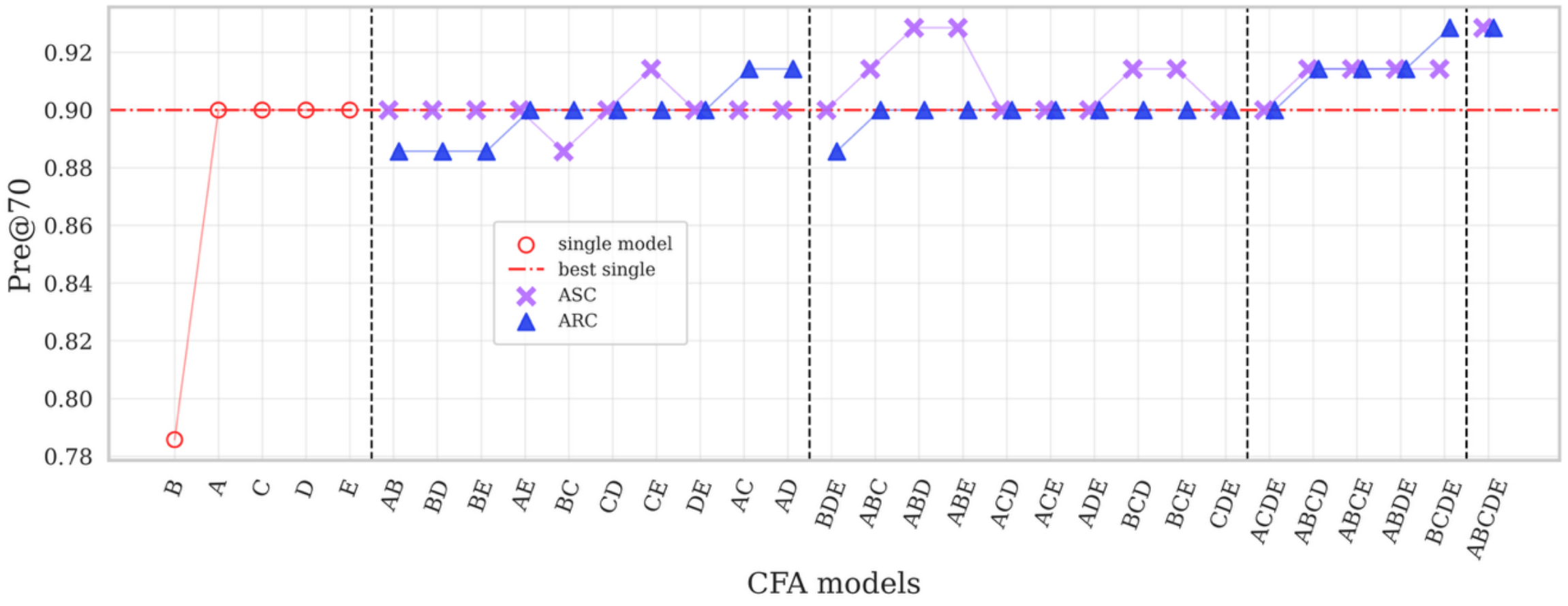

**Fig. 6:** Performance plot for average combination (ASC, ARC) in the **second layer**.

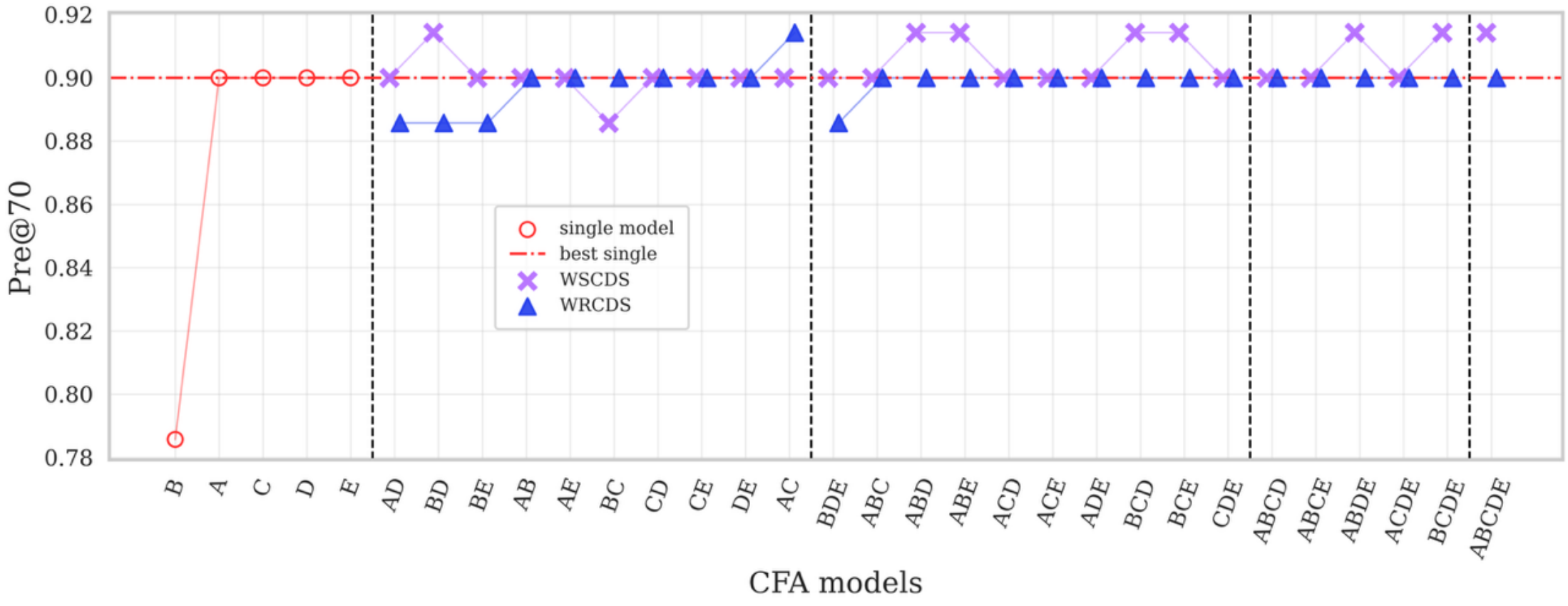

**Fig. 7:** Performance plot for weighted combination (WSCDS, WRCDS) in the **second layer**.

| | Model | Combination Type | Pre@70 |
|---|---|---|---|
| Base model | A | - | **0.9000** |
| | B | - | 0.7429 |
| | C | - | **0.9000** |
| | D | - | 0.8857 |
| | E | - | **0.9000** |
| First layer MCF | ADE | ASC | **0.9143** |
| | ACDE | WSCDS | |
| Second layer MCF | ABD | ASC | **0.9286** |
| | ABE | | |
| | ABCDE | | |
| | BCDE | ARC | |
| | ABCDE | | |

**Table 2:** Performance comparison of base models and best models from multi-layer combinatorial fusion (MCF) across two layers.

Figure 7 presents the performance for weighed combination in the second layer, where many combined models exceed the best base performance of 0.9. Specially, nine models from the WSCDS approach and two models from the WRCDS approach achieve a performance of 0.9143. These results validate the benefit of using diversity strength as a fusion weight. Figure 7 also suggests that score-based weighted fusion is more effective than its rank-based counterpart in this layer. Finally, because the rank-score function graph for the third layer in Figure 4 reveals nearly no diversity, further combinations are not implemented, as they are almost impossible to yield gains.

A summary of base models and the best combined models across 2 layers is in Table 2. The best initial single model achieves a performance of 0.9. After applying MCF in the first layer, the best performance increases to 0.9143. And the second layer further improves this performance to 0.9286. These findings highlight the MCF framework can improve CFA results by stacking multiple fusion layers, effectively exploiting diversity among models to achieve better performance.

# CONCLUSION

Identifying consumer preference is critical in marketing because it helps explain why people choose one product over another. Instead of examining preference through traditional methods such as surveys, ratings, or purchase records, this paper investigates preference detection through eye movement behavior. We use five diverse machine learning models to analyze gaze-based features extracted from an eye-tracking preference dataset. These models are then combined iteratively using the Multi-layer Combinatorial Fusion framework, which refines the model set layer by layer by expanding model combinations and selecting diverse models for the next layer. The results show that MCF proceeds to the second layer and achieves a best Pre@70 of 0.9286, which is higher than both the best first-layer performance and the best original base-model performance. This demonstrates that MCF is an effective framework for improving preference detection and may help predict consumer preference more accurately in marketing applications. In future work, we plan to apply this approach to more realistic consumer-choice settings, such as product images or online shopping interfaces, to further evaluate its practical value.